# COMPOSITIONALITY, SYNONYMY, AND THE SYSTEMATIC REPRESENTATION OF MEANING

**Shalom Lappin** and **Wlodek Zadrozny**
King's College London      IBM T.J. Watson Research Center
shalom.lappin@kcl.ac.uk    wlodz@us.ibm.com

## 1. INTRODUCTION

In a recent issue of *Linguistics and Philosophy* Kasmi and Pelletier (1998) (K&P), and Westerståhl (1998) criticize Zadrozny's (1994) argument that any semantics can be represented compositionally. The argument is based upon Zadrozny's theorem that every meaning function $m$ can be encoded by a function $\mu$ such that (i) for any expression E of a specified language L, $m$(E) can be recovered from $\mu$(E), and (ii) $\mu$ is a homomorphism from the syntactic structures of L to interpretations of L. In both cases, the primary motivation for the objections brought against Zadrozny's argument is the view that his encoding of the original meaning function does not properly reflect the synonymy relations posited for the language.[1]

In this paper we will look at both the technical issues that K&P and Westerståhl raise, and the relation between synonymy and compositionality. We will argue that their technical criticisms do not go through, while the objection from synonymy assumes an additional constraint on compositionality, which they do not make explicit. We also correct some misconceptions about the function $\mu$, e.g. Janssen (1997). We prove that $\mu$ properly encodes synonymy relations, i.e. if two expressions are synonymous, then their compositional meanings are identical. In Sections 2 and 3 we take up K&P's and Westerståhl's formal objections. Section

4 considers the relation between compositionality and the systematic representation of meaning. We suggest that the reason that semanticists have been anxious to preserve compositionality as a significant constraint on semantic theory is that it has been mistakenly regarded as a condition that must be satisfied by any theory that sustains a systematic connection between the meaning of an expression and the meanings of its parts. Recent developments in formal and computational semantics show that systematic theories of meanings need not be compositional.

Zadrozny's argument is based on the observation that a function is defined as a collection of pairs <argument, value>, such that each argument has only one value in the collection (in contrast to a relation in which an argument can be associated with more than one value). The classical version of compositionality is defined as a functional dependence of the meaning of an expression on the meaning of its parts. Since many semantic theories represent meanings as functions, the requirement of compositionality can be expressed as a condition on the <argument, value> pairs that define the function which assigns meanings to the expressions of a language. Specifically, $\mu(s.t) = \mu(\mu(s),\mu(t))$ means that $\mu$ must have among its elements the pairs <s.t, $\mu(\mu(s),\mu(t))$>, <s,$\mu(s)$>, and <t,$\mu(t)$>. Using this approach, the following theorem is proven.

THEOREM 1: Let $M$ be an arbitrary set. Let $A$ be an arbitrary alphabet. Let '.' be a binary operation (concatenation), and let $S$ be the set closure of $A$ under '.'. Let $m: S \rightarrow M$ be an arbitrary function. Then there is a set of functions $M^*$ and a unique map $\mu: S \rightarrow M^*$ such that for all s,t in $S$, $\mu(s.t) = \mu(\mu(s),\mu(t)) = \mu(s)(\mu(t))$, and $\mu(\mu(s)) = m(s)$.

---

[1]Janssen (1997) makes a similar claim.

Theorem 1 entails that even if for the original meaning function $m$ of a language L, $m(A) = m(B)$ and $m(C.A) \neq m(C.B)$, there is a function $\mu$ such that $\mu(C.A) = \mu(C)(\mu(A))$, and $\mu(C.B) = \mu(C)(\mu(B))$.[2]

In the same paper, it is shown (Corollary 2) that the language L doesn't have to be closed under concatenation, and (Proposition 3) that the original meanings given by $m(s)$ can be recovered in other ways, e.g. by having $\mu(s)(\$) = m(s)$, for some element $\$$. We will use the last fact (and method) in the proof of the Synonymy Theorem in Section 3.

## 2. FUNCTIONALITY, TYPE RAISING and K&P's OBJECTIONS

Kasmi and Pelletier (1998) (K&P) raise two main objections. First, they observe that Zadrozny's encoding of L's semantics substitutes functions for the original meanings that $m$ assigns to the expressions of L. So, for example, $\mu(A) = f_1$ rather than $m(A)$, and $\mu(B) = f_2$ rather than $m(B)$, where $m(A) = m(B)$. They conclude that $\mu$ is not a meaning function in the sense that $m$ is. Therefore, while $\mu$ is indeed compositional, it does not encode the semantics of L.

There are two points to make in reply to this criticism. First, K&P cannot object to $\mu$ solely on the grounds that it raises the semantic value of an expression to a function, as such type raising has been standard procedure in formal semantics at least since Montague (1974). In generalized

---

[2]For purposes of simplicity Zadrozny (1994) assumes that the only syntactic operation of L is concatenation, which generates complex expressions like C.A. The argument generalizes to extensions of L that contain additional syntactic operations.

quantifier theory proper names, as well as quantified NP's are taken to denote sets of sets (functions from sets to truth-values) rather than individuals.

Second, while in general $\mu(E) \neq m(E)$, it is always possible to recover $m(E)$ from the function $f$ that $\mu$ assigns to E because $f(E) = m(E)$. Therefore, $\mu$ does, in fact, provide an entirely systematic encoding of the intended interpretation of L as specified by $m$. Of course it is not the case that $\mu$ assigns the same values to the expressions of L that $m$ does. How could it without being identical to $m$? Abstracting away from the fact that $\mu(E)$ is a function that effectively type raises $m(E)$, K&P's objection seems to reduce to the observation that $\mu \neq m$, and so it is not obvious what force attaches to this objection.

K&P's second objection is that it is possible to replace $\mu$ by the identity function $i$ that maps each expression of L to itself without affecting the content of Zadrozny's argument. As $i(E) = E$ and $m(E)$ is the original semantic value of E, it is possible to recover the intended interpretation of E from $i(E)$. Moreover, $i$ is compositional, as $i(C.A) = i(C).i(A)$, and similarly for $i(C.B)$. As $i$ is a trivial encoding of the semantics of L which cannot be taken to be a representation of L's meaning, the same result follows for $\mu$.

The problem with this argument is that its premise is false. While K&P are right to claim that $i$ is a not an encoding of the semantics of E, it is not possible to treat $i$ and $\mu$ as the same type of function here. For any E, $i(E)$ is an argument of $m$ while $\mu(E)$ is another function $f$ such that $f(E) = m(E)$. $i$ does not define a map to the meanings of the expressions in L. It merely yields the domain of $m$, that is the language the semantics of which we are trying to define. The function $i$ in effect says that "the meaning of a sentence is its words". In particular $i$ does not map

synonymous expressions into the same meanings. By contrast, $\mu$ provides a set of functions that assign the appropriate meanings to the expressions of L. Also, as we shall seen in the next section, it maps synonymous expressions into identical meanings. Hence, $\mu$ is not directly parasitic on *m* in the way that *i* is. The functions in $\mu$'s range specify values that correspond to those of *m* for each E. By contrast, *i*(E) does not assign any semantic value to E.

## 3. SYNONYMY, HYPERSETS, AND WESTERSTÅL'S AND JANSSEN'S OBJECTIONS

Westerståhl brings two main arguments against Zadrozny (1994). First, he claims that the use of hypersets in the proof of the theorem is unnecessary because it is possible to reformulate the proof in terms of well founded sets. Specifically, he claims that (i) "Anything that can be described using hypersets can, in principle, be described using only well-founded sets" (p.640), and (ii) concerning the encoding of *m* through a function $\mu$ corresponding to a hyperset, "almost the same enumeration idea can be realized with ordinary sets".

The reply to (i) is as follows. Consider a function that takes itself as an argument, e.g. $f(f) = 27$, and is undefined otherwise. Clearly, a well founded description of *f*, call it *wf*, cannot satisfy $wf(wf) = 27$. The object *wf* is not a function but a well founded graph of a non-well founded function. The justification for using non-well founded sets lies in the need to model cases of self application such as $\mu(a.a) = \mu(a)(\mu(a))$.

To support (ii) Westerstahl constructs the function *v* such $v(a) = \{<<a,0>, m(a)>\} \cup \{<<b,1>, m(a.b)>: b, a.b \text{ in } A\}$. Clearly, *m*(a) is recoverable from *v*(a), and it is possible to

specify an operation APP* where $v(a.b) = APP*(v(a), v(b))$. As, he points out, APP* is intuitively analogous to functional application. However, it is not identical to it. Therefore, Westerstahl has specified a case of pseudo-functional application rather than the real thing. Therefore, he has not actually reconstructed functional application with well founded sets.

Westersahl goes on to point out that function application is not the only semantic operation or the only procedure for developing a theory of meaning. This is certainly correct. However, it is important to note that function application (the construction of complex meanings through functional composition) provides one of the strongest instances of a compositional encoding of meaning. It is more powerful than the mere requirement of a homomorphism between syntactic structure and semantic values. Zadrozny's theorems establish that even if we insist on casting our theory of meaning in functional form, compositionality without additional constraints is, in itself, a vacuous requirement. Hence, the conclusion holds for the weaker condition of homomorphism.

Let us now take up the issue of synonymy. Westerståhl claims that Zadrozny's encoding of *m* through *μ* sustains compositionality at the cost of violating the meaning relations posited for L. Therefore, the semantics which *μ* provides for L is arbitrary. K&P's first objection is a version of this point. Janssen (1997) also raises it:

On the semantic side some doubts can be raised. The given original meanings are encoded using non-well founded sets. It is strange that synonymous sentences get different meanings. (p. 456).

Janssen's objections are misplaced, and they can be addressed through a technical argument. Namely, the function $\mu$ can have the synonymy property, i.e. it can map synonymous expressions into the same meanings. In fact, the function used in the proof of Proposition 3 in Zadrozny(1994) does have the synonymy property.

**Theorem (The synonymy theorem)**: Let $S$ be a language with a binary operation '.', and let $M$ be an arbitrary set. Let $m: S \to M$ be an arbitrary function (assigning meanings to expressions of S). Let $S^*$ be the set $\{ s.\$ : s \in S \}$, where $ is a distinguished element outside S (standing for "end-of-expression").

Then there is a set of functions $M^*$ and a unique map $\mu: S^* \to M^*$ such that for all $s,t \in S$,

1. Compositionality: $\mu(s.t) = \mu(\mu(s),\mu(t))$,
2. Recoverability of meanings: $\mu(s.\$) = m(s)$
3. Synonymy:

    If a and b are synonyms, i.e. they appear in the same strings of S, and can be intersubstituted without the change of meanings; i.e. m(a) = m(b) and m(xay) = m(xby) for all xay in S. Then

    $\mu(a) = \mu(b)$.

For the proof, see the Appendix.

Westerståhl makes his claims on the basis of the fact that where $m(A) = m(B)$,

$\mu(A) \neq \mu(B)$, when $m(A.C) \neq m(B.C)$ for some C. That is, compositionality breaks for C. In fact, that $\mu(A) \neq \mu(B)$ holds, even though $m(A) = m(B)$, is not at all a symptom of ad hoc design for $\mu$. It is an inescapable consequence of (i) the compositionality of $\mu$, and (ii) the fact that $\mu$ must encode the constraint expressed by $m(C.A) \neq m(C.B)$. Clearly, if A and B are not inter-substitutable in all expressions in which they occur as consituents, it is not possible for a compositional function to assign them the same values.[3] As we have seen above, if A and B are inter-substitutable we do in fact have $\mu(A) = \mu(B)$.

It is important to distinguish *two notions of synonymy*. On the first, synonymy corresponds to sameness of meaning in some pre-theoretical sense. On the second, two expressions are synonymous iff they are inter-substitutable in all expressions in which they occur as constituents. We can take *m* as representing the given meaning relations of L. It is reasonable to require that any encoding of *m* that purports to offer an adequate representation of L's semantics preserve, at some level of representation, the meaning relations that *m* defines for L. In fact, $\mu$ does this by virtue of the fact that it assigns functions to the expressions of L that retrieve *m*(E) for each E of L. Specifically, $\mu$ preserves the sameness of meaning that *m* establishes for A and B to the extent that $f_1$ assigns the same value to A that $f_2$ assigns to B. However, if *m* does not establish synonymy between A and B in the sense of inter-substitutivity, this lack of synonymy between A and B is explicitly represented on $\mu$, which assigns them distinct values.

To insist that a compositional semantic function directly encode sameness of meaning taken in the *pre-theoretical* sense (i.e. sameness of value as established either by another meaning

---

[3] See Mates (1952) for a discussion of the relation between compositionality and synonymy in the

function or by linguistic intuition) as identity of value is to add a requirement to the condition of compositionality that goes beyond both the homomorphism constraint and the demand that the function express the given meaning relations of L. Clearly $\mu$ does not satisfy this additional constraint, nor could any compositional function, given the assumptions concerning the meaning relations that *m* defines for L. But this constraint is not explicitly incorporated into the notion of compositionality that semantic theorists have employed, and that K&P and Westerståhl invoke in their criticisms. If it is adopted, then compositionality becomes a stronger condition than the requirement that the meaning of an expression be functionally determined by its structure and the meanings of its constituents.[4] Furthermore, if this pre-theoretical sense of synonymy is crisply defined, we can expect a suitably defined compositional function $\mu$ to reflect it.

## 4. BEYOND COMPOSITIONALITY: SYSTEMATIC RELATIONAL THEORIES OF MEANING

The importance which has been assigned to compositionality in semantic theory can be attributed to the fact that, until recently, many (most?) semanticists have identified it directly

---

sense of inter-substitutivity.

[4] Fulop and Keenan (forthcoming) (F&K) show that the standard formulation of compositionality as the homomorphism requirement on the mapping from the set of syntactic structures of a language to the set of its semantic values is not strong enough to exclude certain "pathological" meaning functions that yield bizarre patterns of interpretation. These patterns are generated by non-standard changes in interpretation across the set of possible models. F&K suggest two successively stronger definitions of compositionality to filter out the problematic meaning functions that they identify. They achieve this filtering effect by incorporating additional constraints upon the set possible models into the definition of a compositional meaning function. While F&K's counterexamples to the standard compositionality condition provide important insight into the limitations of this condition, they do not affect Zadrozny's theorem. As the encoding of a meaning function *m* provided by $\mu$ does not depend upon properties of models, $\mu$ satisfies F&P's strengthened definitions of compositionality.

with the possibility of a systematic representation of meaning. Davidson (1984) appears to go so far as to take compositionality, together with the existence of a finite set of semantic primitives, to be a necessary condition for the learnability of a language. He expresses this idea as follows.

When we regard the meaning of each sentence as a function of a finite number of features of the sentence, we have an insight not only into what there is to be learned; we also understand how an infinite aptitude can be encompassed by finite accomplishments. For suppose that a language lacks this feature; then no matter how many sentences a would-be speaker learns to produce and understand, there will remain others whose meanings are not given by rules already mastered. It is natural to say such a language is *unlearnable*. [italics Davidson's] (p. 8).

It is important to note that in order to sustain a homomorphism from the syntax to the semantics of the sort required by compositionality, a meaning function must apply to expressions with fully specified syntactic representations and yield unique semantic values. Therefore, syntactic and semantic ambiguity are eliminated from the mapping which such a function defines. Ambiguous lexical items are separated into words that stand in a one-to-one correspondence with the distinct senses of the original terms**.** Similarly, the different scope readings of a phrase are obtained from distinct syntactic sources. So, for example, in Montague (1974) a meaning function applies to fully disambiguated syntactic structures in which each lexical item has a unique semantic value, and the relative scope relations of the constituents in a phrase P are fully determined by P's syntactic derivation tree.

In fact, it is possible to construct a theory of meaning which is both non-compositional and systematic.[5] This involves taking the meaning of a syntactically complex expression E to be the set of values of a relation on the meanings of E's syntactic constituents rather than the value of a function. We can sketch the form of such a theory. Let S be the semantic principles (constraints, rules, etc.) of the language L. For each lexical item E of L, S assigns E a set of semantic values. For each syntactically well formed expression E of L with immediate syntactic constituents $e_1,...,e_k$, S generates a relation $R(<sv(e_1),...,sv(e_k)>,int(E))$, where $sv(e_i)$ ($1 \leq i \leq k$) is a set of semantic values, and $sv(E) = \{int(E): R(<sv(e_1),...,sv(e_k)>,int(E))\}$.

A theory of this kind is systematic because for each complex expression E in L, there is a relation $R$ that maps the semantic values of the immediate constituents of E into an interpretation of E $int(E)$, and the semantic value of E is the set $int(E)$ of $R$'s values. Therefore, the meaning of E is determined by the meanings of its constituents. However, as $R$ is not, in general, a function, it maps an ordered k-tuple of meanings into a set of possible interpretations rather than into a single disambiguated interpretation. This set is, in effect, the disjunction of interpretations which can be assigned to E in particular contexts. A relational theory of meaning represents the interpretation of a sentence as underspecified to the extent that it defines its semantic value as a set of possibly incompatible alternative interpretations.

An early instance of such a relational theory of meaning is Cooper's (1983) use of quantifier storage to associate alternative quantifier scope readings with a single syntactic structure. In the

---

[5]See Nerbonne (1996) for a non-compositional approach to semantics in a constraint-based framework that focuses on lexical and syntactic ambiguity. Lappin (forthcoming) gives an overview of non-compositional representations of scope.

computational model of storage presented in Pereira and Shieber (1987) and Pereira (1990) the syntactic structure ST of a sentence which contains quantified NP's is assigned a set of scope interpretations where each element of this set imposes a relative scope ordering on the semantic representations of the quantifiers in ST.

Recent work in underspecified semantics, such as Reyle (1993), Copestake et al. (1995), Crouch and van Genabith (1997), Richter and Sailer (1997), Lappin (1999), and Pollard (1999) extends and generalizes this approach. In each case, the meaning of a sentence is a set of possible interpretations I such that each element of I is obtained by (i) ordering the scope relations of scope taking elements (quantifiers, modals, adverbs, certain connectives, etc.), and/or (ii) assigning values to parameters in the semantic representations of constituents of the sentence. The elements of I are the resolved interpretations of the sentence that are selected in conjunction with information supplied by the discourse context. For each syntactically complex constituent C of a sentence, the constraints in the semantic theory determine the elements of C's semantic value $sv(C)$ by defining the relation $R$ which maps the semantic values of C's immediate constituents into the possible interpretations of C.

## 5. CONCLUSION

We have argued that the objections raised by both K&P and Westerståhl to Zadrozny's theorem concerning compositionality do not hold. Specifically, the main criticism to the effect that the encoding function $\mu$ on which the theorem depends does not preserve the synonymy relations in L specified by the original meaning function $m$ implies an additional constraint on

compositionality that is not part of the original homomorphism requirement. We have also shown that the construction of the compositional function used in Zadrozny(1994) can preserve synonymy relations if that is desired.

Recent work in underspecified semantics has shown that compositionality is not a necessary condition for a systematic semantic theory. This work indicates that it is possible to develop a relational theory of meaning which does not satisfy the homomorphism constraint but does assign semantic values to phrases that are systematically computed from the meanings of their syntactic components. Given this result, there is no reason to take compositionality to be a condition of adequacy on a semantic theory. Therefore, the fact that any meaning function for a language L can be encoded by a mapping from the syntax to the semantics of L that is a homomorphism is not a particularly dramatic result. Even if one adds constraints to the definition of compositionality to exclude functional encodings of the sort employed in the proof of Zadrozny's theorem, such a strengthened condition is of limited interest, given that we can dispense with the homomorphism condition on semantic theory.

APPENDIX: A proof of the synonymy theorem

**Synonymy Theorem**: Let *L* be a language with a binary operation '.', and let *M* be an arbitrary set. Let *m*: S → M be an arbitrary function (assigning meanings to expressions of *L*). Let *L*\* be the set { s.$ : s ∈ L }, where $ is a distinguished element outside L (standing for "end-of-expression").

Then there is a set of functions *M*\* and a unique map μ: L\* → M\* such that

for all s,t ∈ L,

1. Compositionality: μ(s.t) = μ(μ(s),μ(t)),
2. Recoverability of meanings: μ(s.$) = *m*(s)
3. Synonymy:

    If a and b are synonyms, i.e. they appear in the same strings of L, and can be intersubstituted without the change of meanings; i.e. *m(a) = m(b)* and *m(xay)=m(xby)* for all *xay* in *L*. Then

    μ(a) = μ(b).

Proof: We use the Solution Lemma, which says that any set of equations has a unique solution. Thus, in particular if *L1 M* and *L2 M* are solutions to the same set of equations, *L1* and *L2* must be equal. For instance, the system of equations

$X1 = \{p, X1, \{X2\}\}$

$X2 = \{p, X2, \{X1\}\}$

$X3 = \{p, q, X2, \{X1\}\}$

has a unique solution, say $X1= l1$, $X2=l2$, $X3=l3$. Since by renaming the variables $X1$ and $X2$ we obtain an identical set of equations, the assignment $X1=l2$, $X2=l1$, $X3=l3$ is also a solution. It follows that $l1=l2$. This is the idea of the proof. The equations are however slightly more complex.

We should concentrate on synonymy, because compositionality and recoverability follow from the theorems proved in Zadrozny 1994. This proof is an extension of the proofs presented there. Consider $\mu(a)$ and $\mu(b)$. If they are defined as in Proposition 3, Zadrozny 1994, they have the following properties:

$$\mu(a) = \{ <\$,m(a)> \} \cup \{ <\mu(t), \mu(a.t)> : a.t \in L \}$$

$$\mu(b) = \{ <\$,m(b)> \} \cup \{ <\mu(t), \mu(b.t)> : b.t \in L \}.$$

The symbol $ stands for an element that can be used to uniformly decode the original meanings m(s). For completeness, we posit $\mu(\$) = \$$. This property, with an identical condition on other elements of L defines our function $\mu$. That is the idea behind the proof.

To prove the theorem we begin by formulating a set of equations

$$Xa = \{ <X\$,m(a)> \} \cup \{ <Xt, Xa.t : a.t \in L \}$$

$$Xb = \{ <X\$,m(b)> \} \cup \{ <Xt, Xb.t : a.t \in L \}$$

$$X\$ = \{<X\$,X\$> \}$$

which together with all other equations

$$Xs = \{ <X\$,m(s)> \} \cup \{ <Xt, Xs.t : s.t \in L \}, \text{ s in L}$$

$$Xs\$ = m(s), \text{ for s in L}$$

will define a certain set μ as a unique solution to this set of equations (by the solution lemma).

The value of variable Xs is μ(s), for s in L, and the value of the variable X$ is μ($).

So far, the proof follows the line of Zadrozny 1994.

We now use the synonymy property of a and b and make two observations:

1. $\{ <X\$,m(a)> \} = \{ <X\$,m(b)> \}$,

2. Let x and y be (possibly empty) strings in the alphabet of L. Then, if xay is in L, also xby must be in L, since a can be substituted for b (and vice versa). Since these substitutions do not change the meanings, we also have m(xay) = m(xby) for all xay in L.

Substituting p for m(a), and using the fact that a.t is in L if and only if b.t is in L, we get the following (possibly infinite) set of equations

$$Xa = \{ <X\$,p> \} \cup \{ <Xt, Xa.t> : a.t \in L \}$$

$$Xb = \{ <X\$,p> \} \cup \{ <Xt, Xa.t> : a.t \in L \}$$

$$X\$ = \{<X\$,X\$>\}$$

$$Xs = \{ <X\$,m(s)> \} \cup \{ <Xt, Xs.t : s.t \in L \}, \text{ s in } L\setminus\{a,b\}.$$

$$Xs\$ = m(s) \text{ for s in } L$$

The assignement Xa=μ(a), Xb=μ(b), Xs=μ(s) for s in L\{a,b}, together with X$=μ($), Xs$=μ(s.$)=m(s), for all s in L. is a solution (by the definition of the meaning function μ). However, Xb=μ(a), Xa=μ(b), Xs=μ(s) for s in L\{a,b} also produces a solution. Thus, by its uniqueness, μ(a)=μ(b).

As in Zadrozny 1994 we check that compositionality holds, $\mu(s.t) = \mu(s)(\mu(t))$, that is, it follows immediately from the fact that $<Xt, Xs.t>$ is in Xs. Finally, the recoverability of meanings $\mu(s.\$) = m(s)$ holds true, because $\mu(s.\$) = m(s) = \mu(s)(\mu(\$))$, again by direct lookup.

QED.

**Remark 1.** $ doesn't have to extend the language (as it did in Proposition 3). We can use any element which does not make a distinction between the meaning of a and b, and which is outside of the range of *m*. For instance, without extending the language by $ we could recover the meanings by requiring $\mu(s)(\mu) = m(s)$. However, we could not use

$Xa = \{ <a,m(a)> \} \cup \{ <Xt, Xa.t : a.t \in L \}$

$Xb = \{ <b,m(b)> \} \cup \{ <Xt, Xb.t : a.t \in L \}$

because this (or similar construction) would encode the *orthographic difference* between a and b as part of their meanings – which would effectively deny the synonymy of a and b.

**Remark 2.** $\mu(\$)$ is Omega, the simplest non-well founded set, i.e. the canonical set that satisfies X={X}. (cf. Aczel, 1988 p.6).